\documentclass[journal]{IEEEtran}

\usepackage{multirow} 
\usepackage{float} 
\usepackage{makecell}
\usepackage{booktabs}

\usepackage{enumitem}

\usepackage{colortbl} %
\definecolor{gray}{rgb}{ .851,  .851,  .851}

\usepackage{amsmath,amsfonts}
\usepackage{algorithmic}
\usepackage{algorithm}
\usepackage{array}
\usepackage[caption=false,font=normalsize,labelfont=sf,textfont=sf]{subfig}
\usepackage{textcomp}
\usepackage{stfloats}
\usepackage{url}
\usepackage{verbatim}
\usepackage{graphicx}
\usepackage{cite}
\begin{document}

\title{Dual-supervised Asymmetric Co-training for Semi-supervised Medical Domain Generalization}

\author{Jincai Song, Haipeng Chen, Jun Qin, Na Zhao
\thanks{This work was supported by the National Natural Science Foundation of China (62276112) and the Jilin Province Science and Technology Development Plan Key R\&D Project (20230201088GX). \textit{(Corresponding authors: Jun Qin; Na Zhao.)}}
\thanks{Jincai Song and Haipeng Chen are with the College of Computer Science and Technology, Jilin University, Changchun, Jilin 130012, China, and with the Key Laboratory of Symbolic Computation and Knowledge Engineering of the Ministry of Education, Jilin University, Changchun, Jilin 130012, China (e-mail: songjc21@mails.jlu.edu.cn; chenhp@jlu.edu.cn).}
\thanks{Jun Qin is with the School of Computer Science and Technology, Changchun University of Science and Technology, Changchun 130012 (e-mail: qinjun@cust.edu.cn).}
\thanks{Na Zhao is with the Information Systems Technology and Design Pillar, Singapore University of Technology and Design, Singapore 487372 (e-mail: na$\_$zhao@sutd.edu.sg).}
}

\markboth{IEEE TRANSACTIONS ON MULTIMEDIA,~Vol.~14, No.~8, August~2021}%
{Shell \MakeLowercase{\textit{et al.}}: A Sample Article Using IEEEtran.cls for IEEE Journals}

\IEEEpubid{0000--0000/00\$00.00~\copyright~2021 IEEE}

\maketitle

\begin{abstract}
Semi-supervised domain generalization (SSDG) in medical image segmentation offers a promising solution for generalizing to unseen domains during testing, addressing domain shift challenges and minimizing annotation costs. However, conventional SSDG methods assume labeled and unlabeled data are available for each source domain in the training set, a condition that is not always met in practice. 
The coexistence of limited annotation and domain shift in the training set is a prevalent issue.
Thus, this paper explores a more practical and challenging scenario, cross-domain semi-supervised domain generalization (CD-SSDG), where domain shifts occur between labeled and unlabeled training data, in addition to shifts between training and testing sets. 
Existing SSDG methods exhibit sub-optimal performance under such domain shifts because of inaccurate pseudo-labels.
To address this issue, we propose a novel dual-supervised asymmetric co-training (DAC) framework tailored for CD-SSDG. 
Building upon the co-training paradigm with two sub-models offering cross pseudo supervision, our DAC framework integrates extra feature-level supervision and asymmetric auxiliary tasks for each sub-model.
This feature-level supervision serves to address inaccurate pseudo supervision caused by domain shifts between labeled and unlabeled data, utilizing complementary supervision from the rich feature space. 
Additionally, two distinct auxiliary self-supervised tasks are integrated into each sub-model to enhance domain-invariant discriminative feature learning and prevent model collapse.
Extensive experiments on real-world medical image segmentation datasets, \textit{i.e.}, Fundus, Polyp, and SCGM, demonstrate the robust generalizability of the proposed DAC framework.
\end{abstract}


\begin{figure}[t]
  \centering
   \includegraphics[width=1\linewidth]{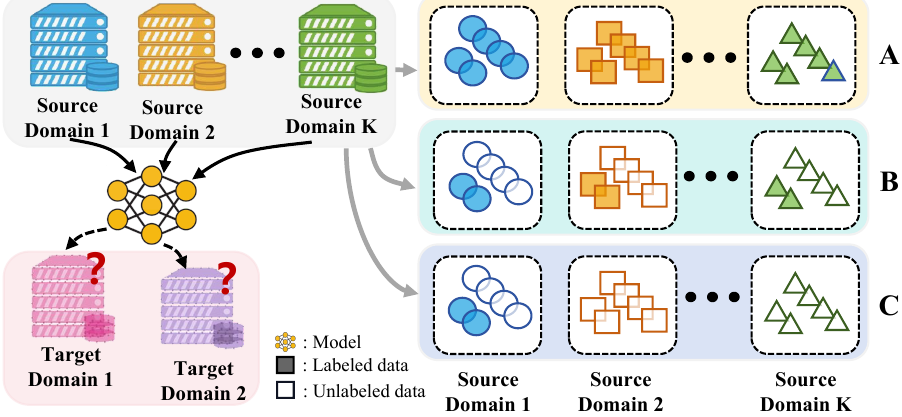}
   \caption{\small{
   \textbf{CD-SSDG} (C) is closely related to DG (A) and SSDG (B), but poses unique challenges that cannot be solved by DG or SSDG methods alone, \textit{i.e.},  partially-labeled training data and domain shifts between labeled and unlabeled training data.}
   }
   \label{fig:teaser}
\end{figure}

\section{Introduction}
\label{sec:intro}

\IEEEPARstart{M}{edical} image segmentation, a critical task for computer-aided diagnosis and surgery, has advanced significantly in the era of deep learning \cite{yang2023directional,chen2023adaptive,liu2024causal,zhang2023hmnet}. 
However, most methods are often evaluated under the assumption of an independent and identically distributed (\textit{i.i.d.}) relationship between training and testing sets, which is unrealistic given the prevalent occurrence of distribution shifts in real-world clinical applications.
In medical image segmentation, distribution shifts primarily arise from covariate shifts (\textit{a.k.a.} domain shifts) \cite{muandet2013domain,zhou2022domain,yoon2023domain}  (\textit{e.g.}, variations in brightness and contrast, as shown in Fig.~\ref{fig:dataset}), rather than label shifts \cite{zhou2022domain}, due to the structural similarities within human anatomy \cite{saladin2010anatomy}.
Such shifts can be attributed to external variations in patient demographics \cite{li2020self,wang2020meta} or scanner types \cite{tao2019deep}, whereas internal variability may also occur within a controlled setting (\textit{e.g.}, the same scanner or health care facility) due to factors such as hardware aging \cite{yoon2023domain} or scan acquisition settings \cite{zhang2020generalizing}.
Unfortunately, methods based on the \textit{i.i.d.} assumption experience a performance decrease when faced with domain shifts in unknown testing environments \cite{zhou2022domain}, limiting their application in real-world scenarios. 
Consequently, there is a growing demand for domain generalization (DG) techniques that can adapt to the inherent domain shifts between training and testing sets, especially in the context of medical image segmentation, given the high degree of data heterogeneity \cite{yoon2023domain}.
\IEEEpubidadjcol

Most DG methods in medical image segmentation, despite their adaptability, necessitate full supervision from multiple source domains for training. This is impractical because of the cost and feasibility constraints associated with annotating diverse medical images in real-world scenarios, especially considering the need for medical knowledge and clinical experience \cite{gao2023correlation,basak2023pseudo,xie2023is2net,nie2023deep}. 
In response, semi-supervised domain generalization (SSDG) \cite{liu2021semi,liu2022vmfnet,yao2022enhancing} was proposed to address this data-intensive challenge while ensuring generalizability to shifted domains in testing environments. 
Nonetheless, existing SSDG methods typically assume that labeled data are derived from multiple source domains and that labeled and unlabeled data within each source domain are \textit{i.i.d.} (as depicted in Fig.~\ref{fig:teaser}). This assumption may not hold in practical scenarios. 
During clinical data collection, given a certain amount of labeled data from a typical center, since unlabeled data are easily accessed and collected, unlabeled data may originate from multiple centers. When physicians have a massive amount of unlabeled data, they may lack sufficient time to verify whether the data belong to the same distribution.
In this work, we study a more realistic setting termed cross-domain semi-supervised domain generalization (CD-SSDG), where labeled data originate from \textit{a single source} domain and unlabeled data can originate from domains different from the that of the labeled data, as illustrated in Fig.~\ref{fig:teaser}.

\begin{figure}[t]
  \centering
   \includegraphics[width=1\linewidth]{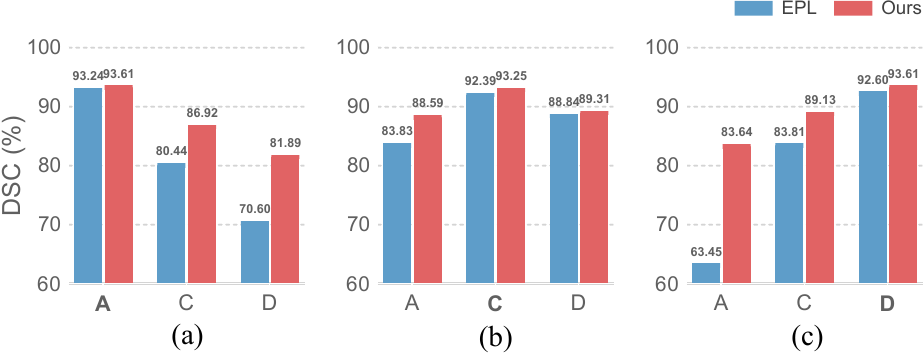}
   \caption{\small{
   {Comparisons of the pseudo-label quality generated by the SSDG method EPL \cite{yao2022enhancing}
 and our DAC under the CD-SSDG scenario with different single-source labeled domain (\textbf{bold}) on the Fundus dataset using $20\%$ labeled data, with domain B as the target domain. }
   }}
   \label{fig:pselabel}
\end{figure}

In the context of CD-SSDG, the coexistence of limited annotations and domain shifts between labeled and unlabeled training data presents significant challenges for existing SSDG methods. These methods face challenges in achieving satisfactory generalizability by applying the information gained from labeled data directly to predict pseudo-labels for many unlabeled medical images across multiple domains. 
These challenges may stem from the negative impact of domain shifts on the pseudo supervision of unlabeled images. 
For example, the quality of pseudo-labels may be poor in an unlabeled domain (\textit{e.g.}, A) that exhibits a substantial domain gap with the single labeled source domain (\textit{e.g.}, D) in Fig.~\ref{fig:pselabel}(c), and these noisy pseudo-labels may also exert a negative effect on domain generalizability learning,
as evident in the results presented in Table~\ref{tab:table1} and Fig.~\ref{fig:Ratio}.

To overcome these limitations, we introduce a novel framework named dual-supervised asymmetric co-training (DAC) for CD-SSDG.
The proposed DAC framework incorporates additional feature-level supervision alongside pseudo supervision based on a co-training scheme \cite{yao2022enhancing}, contributing to a dual supervision mechanism.
Moreover, it introduces different auxiliary tasks for co-trained sub-models, leading to an asymmetric co-training scheme.
Specifically, to alleviate the adverse impacts of inaccurate pseudo-label supervision, we utilize the original features from one sub-model as feature-level supervision for style augmented features of another sub-model, providing correct directions for learning domain-invariant features.
As feature-level supervision encodes rich underlying knowledge, it can serve as a form of soft supervision and supplement class-specific information with generic details that are potentially omitted in pseudo-label supervision commonly used in existing SSDG approaches.

Furthermore, to enhance discriminative feature learning, which can mitigate the impact of noisy pseudo-labels and is inspired by the success of self-supervised learning in the image domain, we choose it as the auxiliary task. 
Specifically, we design an asymmetric co-training scheme by incorporating different self-supervised auxiliary tasks for each sub-model rather than applying a single auxiliary task to both sub-models. This asymmetric scheme encourages the sub-models to learn discriminative yet diverse features, thereby facilitating the learning of more accurate pseudo-labels in the domain-shifted training environment, as shown in Fig.~\ref{fig:pselabel}. 
Moreover, considering the distinctive characteristics of medical images, \textit{e.g.,} varied target sizes and blurred edges in addition to style variations, we opt for localization and rotation as pretext tasks and adapt them to our co-training framework. Specifically, the first self-supervised task involves localizing the mixed region in the CutMix image composed of two styles of augmented images, named mixed patch localization (mixPatchLoc). This aids the first sub-model in learning discriminative features of segmentation target structures. The second self-supervised task involves randomly rotating an arbitrary patch within the style-augmented input image and focuses on predicting the rotation degree of this rotated patch, named random patch rotation prediction (randPatchRot). The randPatchRot enhances the second sub-model's high-level semantic understanding of targets. Notably, the asymmetric scheme is implemented only during the training process, ensuring that there is no impact on the inference complexity. 

\noindent
Our contributions can be summarized as follows:

\begin{itemize}[leftmargin=*]
    \item We propose a novel dual-supervised asymmetric co-training (DAC) framework to address practical and challenging CD-SSDG problems in medical image segmentation tasks, with only a few annotated samples from \textit{a single source domain} used as labeled data.
    \item We introduce a dual supervision mechanism and an asymmetric co-training scheme within the DAC. These two components work together to mitigate the detrimental effects of inaccurate pseudo-labels caused by domain shifts in the training set and foster the learning of domain-invariant discriminative features.
    \item We conduct extensive experiments on three medical image segmentation datasets and demonstrate the superior performance of DAC over other SSDG baselines under various CD-SSDG configurations with different labeled ratios.
\end{itemize}

\section{Related Work}
\label{sec:related_work}

\noindent
\textbf{Domain Generalization.} DG aims to develop a robust model that can generalize to any unseen out-of-distribution target domain with multiple\footnote{A special and extreme case of DG is single domain generalization (sDG), which assumes only one source domain. Owing to space limitations, we omit the discussion on sDG and direct readers to \cite{vidit2023clip,qu2023modality}
for more details.} source domains. To achieve DG, various techniques can be employed, including 1) domain alignment \cite{shao2019multi,zhao2022style,wang2021respecting,zhao2024style}, 2) meta-learning \cite{balaji2018metareg,choi2021meta}, 3) data augmentation \cite{zhou2023mixstyle,10073578,zhao2022synthetic}, and 4) self-supervised learning \cite{moon2022tailoring,10347519,jiao2025domain}.

In medical image segmentation, several methods \cite{liu2021feddg,khandelwal2020domain,liu2020shape,li2022domain} adopt meta-learning to address DG challenges. For example, FedDG \cite{liu2021feddg} employs a meta-update objective function to guide cross-domain optimization while considering the ambiguous boundary area. 
ETTA-SE \cite{li2022domain} introduces an innovative meta-objective by combining episodic training with task augmentation to align across training domains.
In addition, recent methods \cite{zhou2022generalizable,su2023rethinking,gu2023cddsa} employ data augmentation techniques, yielding encouraging results. For instance,
CDDSA \cite{gu2023cddsa} applies disentangled anatomical representations and newly constructed style codes to simulate images with different styles.

\noindent
\textbf{Semi-supervised Semantic Segmentation.} 
Semi-supervised semantic segmentation reduces annotation efforts by using a small amount of labeled data with a larger pool of unlabeled data while aiming to achieve performance comparable to that of fully supervised methods. 
One popular technique for semi-supervised learning is self-training, which iteratively assigns pseudo-labels to unlabeled samples \cite{li2022pseco,ma2023enhanced}.
In contrast to self-training, which depends on the predictions of a single model to generate pseudo-labels, co-training \cite{chen2021semi,wang2023conflict} leverages the collaboration of dual learners to extract more effective information from unlabeled data. 

{
Recently, many efforts \cite{wu2023acl,liu2022act,zhang2024quality,li2022hierarchical,tian2023implicit,zhang2023self,shu2022cross,basak2023pseudo,du2023coarse,wang2023mcf,10739349,bai2023bidirectional} have been made in semi-supervised medical image segmentation. For instance,
ACT \cite{liu2022act} introduces an asymmetric framework by training two segmentors on different labeled data and integrating their knowledge through co-training.
QDC-Net \cite{zhang2024quality} combines a conventional network with an evidential network to promote appropriate prediction divergence for cross-supervised training.
HDUS \cite{li2022hierarchical} introduces a hierarchical deep network with an uncertainty-aware semi-supervised learning framework to address the issue of limited manual annotations.
IEPAlign \cite{tian2023implicit} and SCP-Net \cite{zhang2023self} leverage prototypical learning to increase prediction diversity and increase training efficacy in semi-supervised consistency learning. 
PLGCL \cite{basak2023pseudo} uses pseudo-labels to guide patch-wise contrastive learning to learn discriminative representations. 
CRII-Net \cite{du2023coarse} uses intra-patch ranked loss and inter-patch ranked loss mechanisms to achieve fine-grained segmentation results for challenging regions.
DCPLG \cite{wang2023mcf} employs two different sub-models to address discrepancies between sub-models and correct cognitive biases of the model.
ETC-Net \cite{10739349} introduces an evidence-based tri-branch consistency learning framework to address the challenges of cognitive bias and noise components in pseudo-labels.
BCP \cite{bai2023bidirectional} uses unlabeled data to learn common semantics from labeled data by copy-pasting labeled and unlabeled data bidirectionally.
}

\noindent
\textbf{Semi-supervised Domain Generalization.}
SSDG addresses the challenge of adapting to unseen domains during testing while mitigating the impact of domain shifts and reducing the need for costly annotations. As a recently proposed approach, SSDG has attracted significant attention \cite{zhou2023semi,wang2023better,zhou2023mixstyle}. StyleMatch \cite{zhou2023semi} assigns pseudo-labels to unlabeled data and employs a style transfer model to augment the domain space. BPL \cite{wang2023better} combines domain-aware labels and a dual classifier to produce high-quality pseudo-labels for class tasks, whereas SSGM \cite{li2021semantic} uses a generative model to capture the joint image-label distribution for SSDG in semantic segmentation. 

Several efforts have been made to address SSDG in the context of medical image segmentation.
DGNet \cite{liu2021semi} models domain shifts via disentanglement and extracts robust anatomical features for predicting segmentation results by applying multiple constraints via a gradient-based meta-learning approach. 
vMFNet \cite{liu2022vmfnet} represents the compositional components of human anatomy as learnable von-Mises-Fisher kernels that are robust to cross-center images. 
A\&D \cite{wang2024towards} utilizes a diffusion encoder to extract distribution-invariant features from multiple domains.
$IS^2$Net \cite{xie2023is2net} introduces a semantic and style memory bank to strengthen feature representations within a domain and build cross-domain relationships. SOFR \cite{zhang2022semi} introduces two regularization terms to encourage the model to learn domain-invariant and complete features. EPL \cite{yao2022enhancing} presents a co-training framework with cross pseudo supervision during training for better generalization.

Unfortunately, most existing SSDG medical image segmentation methods typically rely on multi-source domain labeled data, which may not be readily accessible in real-world scenarios. Furthermore, these methods face challenges when domain shifts occur in the training data, resulting in sub-optimal segmentation results due to the shared distribution assumption between labeled and unlabeled data.
To address these limitations, we focus on cross-domain semi-supervised domain generalization (CD-SSDG) and introduce a dual-supervised asymmetric co-training framework to handle domain shifts between labeled and unlabeled data, as well as between training and testing sets.


\begin{figure*}
  \centering
  \includegraphics[width=0.98\textwidth]{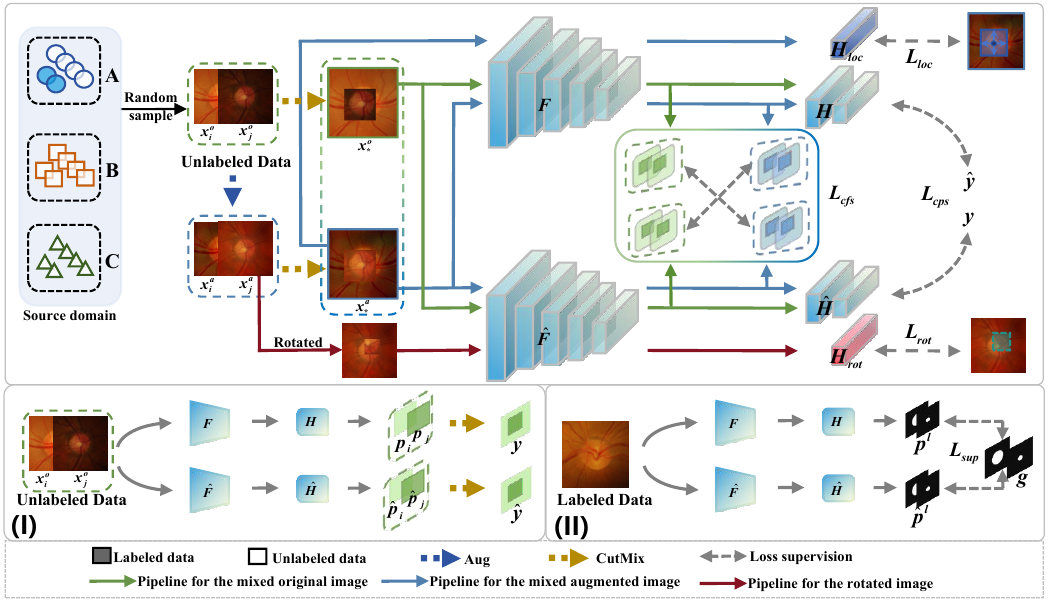}
  \caption{\small{\textbf{Dual-supervised asymmetric co-training (DAC) framework for CD-SSDG}. A, B, and C represent three distinct source domains, with only A containing a limited amount of labeled data. DAC comprises two sub-models, each containing a feature extractor ($F$/$\hat{F}$) and a segmentation head ($H$/$\hat{H}$). The two modules of each sub-model are supervised by \textit{dual-level supervision} ($L_{cps}$ and $L_{cfs}$) provided by the other sub-model using unlabeled data, along with the supervised loss $L_{sup}$ using labeled data. Moreover, each sub-model is equipped with an additional head ($H_{loc}$/$H_{rot}$) to integrate different auxiliary self-supervised learning tasks, mixPatchLoc ($L_{loc}$) and randPatchRot ($L_{rot}$), leading to an \textit{asymmetric co-training} scheme. \textbf{(\MakeUppercase{\romannumeral1}) The pseudo-label generation process} (\textit{cf.} Section~\ref{subsec:DL} for details). \textbf{(\MakeUppercase{\romannumeral2}) Pipeline for labeled data}. In this figure, $p^l$/$\hat{p}^l$ and $g$ represent the predicted probability maps and ground truth masks for the optic disc and cup, respectively.
 }}
  \label{fig:DAC}
\end{figure*}


\section{Proposed Method}
\label{sec:method}

\noindent
\textbf{Problem Definition.} In CD-SSDG, the training data are composed of $K$ similar but distinct source domains. The labeled set from one source domain is denoted as $S^l_1=(X^l,Y^l)$, where $X^l=\{x_i\}^{N^l}_{i=1}$ denotes the input space and $Y^l=\{y_i\}^{N^l}_{i=1}$ represents the label space. $N^l$ denotes the number of labeled samples. Additionally, the unlabeled set from $K$ source domains is denoted as $\{S^u_1,S^u_2,...,S^u_K\}$, where $S^u_k=(X^u_k)$, and $X^u_k=\{x_i\}^{N^u_k}_{i=1}$. $N^u_k$ denotes the number of unlabeled samples from the $k$-th domain. 
Note that the marginal distribution $P_S(X^l)\neq P_S(X^u_k)$ for $k\in \{2, ...,K\}$, and $P_S(X^u_i)\neq P_S(X^u_j)$ for $i,j\in \{1, ...,K\}$, $i\neq j$, but the label space $P_S(Y^l) = P_S(Y^u_k)$ for $k\in \{2, ...,K\}$, and $P_S(Y^u_i) = P_S(Y^u_j)$ for $i,j\in \{1, ...,K\}$, $i\neq j$.
During testing, the model is deployed in an unseen target domain $T=(X)$, where the marginal distribution of the target domain is different from that of any source domain, \textit{i.e.}, $P_T(X)\neq P_S(X^l)$ and $P_T(X)\neq P_S(X^u_k)$ for $k\in \{1, ...,K\}$, while the label space remains the same, \textit{i.e.}, $P_T(Y) = P_S(Y^l)$ and $P_T(Y) = P_S(Y^u_k)$ for $k\in \{1, ...,K\}$.

Our objective is to develop a segmentation model for medical images that can be generalized across domains by effectively utilizing labeled and unlabeled source data, with the presence of domain shifts between them in addition to shifts between the training and testing sets.

\noindent
\textbf{Framework Overview.}
Fig.~\ref{fig:DAC} shows the framework of our dual-supervised asymmetric co-training (DAC) approach. DAC comprises two parallel sub-models that share the same structure but are initialized differently. Each sub-model consists of a feature extractor, denoted as $F$ or $\hat{F}$, and a segmentation head, denoted as $H$ or $\hat{H}$. DAC introduces two novel components: 1) \textbf{dual-supervised co-training} (\textit{cf.} Section~\ref{subsec:DL}), which integrates feature-level supervision with pseudo supervision to learn domain-invariant features, and 2) \textbf{asymmetric co-training} (\textit{cf.} Section~\ref{subsec:AC}), which incorporates different self-supervised auxiliary tasks for each sub-model by adding additional prediction heads, \textit{i.e.}, $H_{loc}$ for mixPatchLoc and $H_{rot}$ for randPatchRot, to enhance discriminative feature learning.

\subsection{Dual-supervised Co-training}
\label{subsec:DL}
In our DAC framework, the outputs of the unlabeled data from each sub-model are supervised via dual-level supervision: \textit{cross pseudo supervision} and \textit{cross feature supervision}, which are generated by the other sub-model.
Cross pseudo supervision provides confident and class-specific information for the regularization process. However, its accuracy can be influenced by the domain shift between labeled and unlabeled data, particularly in challenging segmentation cases such as small or indistinct targets, as illustrated in Fig.~\ref{fig:Q_C_polyp}.  
Moreover, this approach could omit valuable information from unlabeled data associated with low-confidence pseudo-labels. In contrast, cross feature supervision introduces a form of soft regularization by enhancing consistency and supplementing class-specific information with generic details extracted from the rich feature space. 
Both supervision mechanisms contribute to the collaborative learning of domain-invariant features, thereby enhancing the model's generalizability.

During training, we randomly select two images $(x_{i}^o, x_{j}^o)$ from the unlabeled set. To augment our input data, we apply the style augmentation (Aug) method following \cite{yao2022enhancing,xu2021fourier}, which yields augmented images $(x_{i}^a, x_{j}^a)$ that have the same content but a different style from the original images $(x_{i}^o, x_{j}^o)$. Then, we employ CutMix to create the mixed original and augmented image, denoted as: 
\begin{align}
  x^o_* = x_{i}^o\odot \mathbf{M} + x_{j}^o\odot \mathbf{(1-M)}, \\
  x^a_* = x_{i}^a\odot \mathbf{M} + x_{j}^a\odot \mathbf{(1-M)},
  \label{eq:cutmix_img}
\end{align}
where $\mathbf{M}$ is a binary zero-centered mask used to control the spatial range if the image is mixed, $\mathbf{1} \in \{1\}^{H \times W}$, $H$ and $W$ denote the height and width of the input image, respectively, and $\odot$ represents element-wise multiplication. The size of the zero-value region is $\beta H \times \beta W$, where $\beta \sim U(\tau_1,\tau_2)$, and $0< \tau_1 < \tau_2 < 1$.

Subsequently, $(x^o_*, x^a_*)$ are fed into two sub-models, and their probability maps are computed as:
\begin{align}
  p^o_* = H(F(x^o_*)), \quad p^a_* = H(F(x^a_*)),\\
  \hat{p}^o_* = \hat{H}(\hat{F}(x^o_*)), \quad
  \hat{p}^a_* = \hat{H}(\hat{F}(x^a_*)).  
\end{align}
  \label{eq:pred}
\noindent
\textbf{Cross Pseudo Supervision (CPS).} 
Unlike EPL \cite{yao2022enhancing}, we use original images $(x_{i}^o, x_{j}^o)$ in a sub-model to generate supervisory signals for another sub-model to obtain a high-quality pseudo-label.
After obtaining probability maps $p_{(\cdot)}^o$ and $\hat{p}_{(\cdot)}^o$ of the original image $x_{i}^o / x_{j}^o$ from the two sub-models, we apply the same mask $\textbf{M}$ to the input images to generate the pseudo-labels for each sub-model:
\begin{equation}
\begin{gathered}
  y = B_{\sigma} (p_{i}^o\odot \mathbf{M} + p_{j}^o\odot \mathbf{(1-M)}),\\
  \hat{y} = B_{\sigma} (\hat{p}_{i}^o\odot \mathbf{M} + \hat{p}_{j}^o\odot \mathbf{(1-M)}),
  \label{eq:pseudo}
\end{gathered}
\end{equation}
where $B_{\sigma}$ represents operators that convert the mixed probability map into a hard pseudo-label.

To ensure the selection of high-quality pseudo-labels and avoid the inclusion of low-quality labels, 
we use the prediction variance of $x^o_*$ and $x^a_*$ as a confidence measure for pseudo-labels, following the approach of EPL \cite{yao2022enhancing}. 
Specifically, the mean squared error (MSE) between $p^o_*$ and $p^a_*$ is considered the variance $V$. Similarly, $\hat{V}$ is the MSE between $\hat{p}^o_*$ and $\hat{p}^a_*$.
The confidence-aware cross pseudo supervision loss $L_{cps}$ is subsequently computed as:
\begin{equation}
\begin{aligned}
  L_{cps} &=e^{-V}l_{ce}(\hat{p}^o_*, y) + e^{-\hat{V}}l_{ce}(p^o_*, \hat{y}) + V + \hat{V},
\end{aligned}
\label{eq:unloss}
\end{equation}
where $l_{ce}$ represents the cross-entropy loss.

\noindent
\textbf{Cross Feature Supervision (CFS).} In addition to the pseudo-labels, we incorporate feature-level supervision, ensuring that the feature output of the mixed augmented input in one sub-model remains consistent with the feature of the mixed original input from another sub-model, regardless of style augmentation and model differences. This approach encourages the model to learn domain-invariant structural features, thus allowing it to generalize effectively to the unseen domain. Moreover, CFS can provide more effective supervised signals for unlabeled data training, especially for difficult regions that are easily ignored by CPS, facilitating the model in learning stable and generalizable features. Cross feature supervision is formulated as follows:
\begin{equation}
\begin{aligned}
  L_{cfs} = l_{mse}(\hat{F}(x^a_*), F(x^o_*)) + l_{mse}(F(x^a_*), \hat{F}(x^o_*)),
\end{aligned}
\label{eq:CFS}
\end{equation}
where $l_{mse}$ represents the MSE loss.

\subsection{Asymmetric Co-training}
\label{subsec:AC}
Unfortunately, the reliability of dual supervision is not always guaranteed owing to the risk of model collapse. Moreover, the dataset scale for medical image segmentation tasks is typically much smaller than that for natural images because of the high cost of data acquisition, privacy concerns and the labor-intensive nature of manual annotation by medical experts. These limitations constrain the learning of discriminative features for generating accurate pseudo-labels and generalizing them to the unseen testing domain.
To overcome these challenges, we incorporate self-supervised learning as an auxiliary task for sub-models to maximize the exploitation of the limited available data, enhancing the learning of discriminative and transferable features. Instead of applying a single self-supervised learning task to both sub-models, we implement an asymmetric co-training scheme, which involves incorporating different self-supervised auxiliary tasks for each sub-model. The asymmetric scheme serves two main design motivations: 
1) Encouraging the sub-models to learn discriminative yet diverse features, thereby generating more accurate pseudo-labels and enhancing the model's robustness in domain-shifted training environments;
2) Preventing the convergence of the two sub-models into a single representation, thereby ensuring the efficacy of feature-level supervision. 

\begin{figure}[t]
  \centering
   \includegraphics[width=0.98\linewidth]{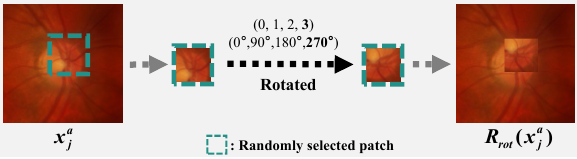}
   \caption{\small{
   \textbf{Generation of a randomly rotated patch}. The rotation degree of the randomly selected image patch is chosen randomly from $\{0^\circ,90^\circ,180^\circ,270^\circ \}$.
   }}
   \label{fig:randPatchRot}
\end{figure}

Specifically, we introduce the first self-supervised task, termed mixed patch localization (mixPatchLoc), where the goal is to localize the mixed patch in the CutMix image $x^a_*$, which consists of two augmented images. This task helps the sub-model learn discriminative features of segmentation target structures in medical images by detecting discontinuities at the border of the mixed patch.
These features contribute to enhancing the sub-model's understanding of the semantics of medical images, including both the original image and the mixed patch.  
Additionally, we introduce the second self-supervised task, termed random patch rotation prediction (randPatchRot), which involves predicting the rotation of a randomly rotated patch within the augmented input image $x_{j}^a$. The objective of randPatchRot is to encourage the sub-model to generate coherent predictions regardless of the random rotation applied to the localized potential target, thereby enhancing its high-level semantic understanding of the segmentation target.
Notably, the inputs to both auxiliary tasks are style-augmented multi-domain unlabeled images, thus enabling the model to identify discriminative domain-invariant features for images with various styles. 
Moreover, both self-supervised tasks are easy to implement and computationally inexpensive, requiring only one linear layer each.

\noindent
\textbf{Mixed patch localization (mixPatchLoc)}. The mixPatchLoc aims to localize the mixed patch, \textit{i.e.}, $x_{j}^a\odot \mathbf{(1-M)}$, within the augmented CutMix image $x^a_*$. We represent the ground-truth location of this mixed patch as $\mathbf{t} = [c_x, c_y, d]$, where $(c_x, c_y)$ denotes the position of the patch center and $d$ is half of the side length. We introduce an additional head $H_{loc}$ to the first sub-model $F$ for predicting the localization of the mixed patch. The predicted localization is compared with the ground truth via the mean absolute error: 
\begin{equation}
  L_{loc} = l_{mae}(H_{loc}(F(x^a_*)),\textbf{t}).
  \label{eq:Dec}
\end{equation}

\noindent
\textbf{Random patch rotation prediction (randPatchRot).} In addition to tasking one sub-model with localizing the unknown mixed patch in the augmented CutMix image, we incorporate another sub-model with a rotation prediction task, a simple yet effective pretext task for self-supervised learning \cite{gidaris2018unsupervised}. The pre-processing approach for rotation prediction is illustrated in Fig.~\ref{fig:randPatchRot}. We utilize the same augmented image $x_{j}^a$ but randomly choose a patch of size $\alpha \beta H \times \alpha \beta W$, where $\alpha \in (0,1)$, instead of using the mixed patch. Subsequently, we rotate the patch by randomly selecting an angle from $\{0^\circ,90^\circ,180^\circ,270^\circ \}$. The random selection of the patch and random rotation prevent the model from easily recognizing the target region via low-level clues, thereby encouraging the learning of discriminative features with an understanding of high-level semantics.
We introduce an additional head $H_{rot}$ to the second sub-model $\hat{F}$ for predicting the rotation degree of the randomly selected patch. The rotation prediction is treated as a classification problem, where the predicted rotation degree is compared with the chosen degree $r$ via the cross-entropy loss: 
\begin{equation}
  L_{rot} = l_{ce}(H_{rot}(\hat{F}(R_{rot}(x_{j}^a)), r)),
  \label{eq:Rot}
\end{equation}
where $R_{rot}$ represents the two pre-processing steps for generating a randomly rotated patch, as detailed in Fig.~\ref{fig:randPatchRot}.

\subsection{Dual-supervised Asymmetric Co-training}
\label{subsec:DAC}
By integrating dual supervised co-training and two self-supervised auxiliary tasks within the asymmetric co-training, our DAC approach provides varied and thorough supervision for unlabeled data. 
Additionally, DAC determines the supervised loss $L_{sup}$ for labeled data using the dice loss. Consequently, the total loss for training DAC is:  
\begin{equation}
\begin{aligned}
  L = L_{sup} + \lambda_{cps} L_{cps} + \lambda_{cfs} L_{cfs} + \lambda_{ac} (L_{loc} + L_{rot}),
\end{aligned}
\label{eq:total_loss}
\end{equation}
where $\lambda_{cps}$, $\lambda_{cfs}$, and $\lambda_{ac}$ are balance coefficients.

\begin{figure}
  \centering
   \includegraphics[width=1.0\linewidth]{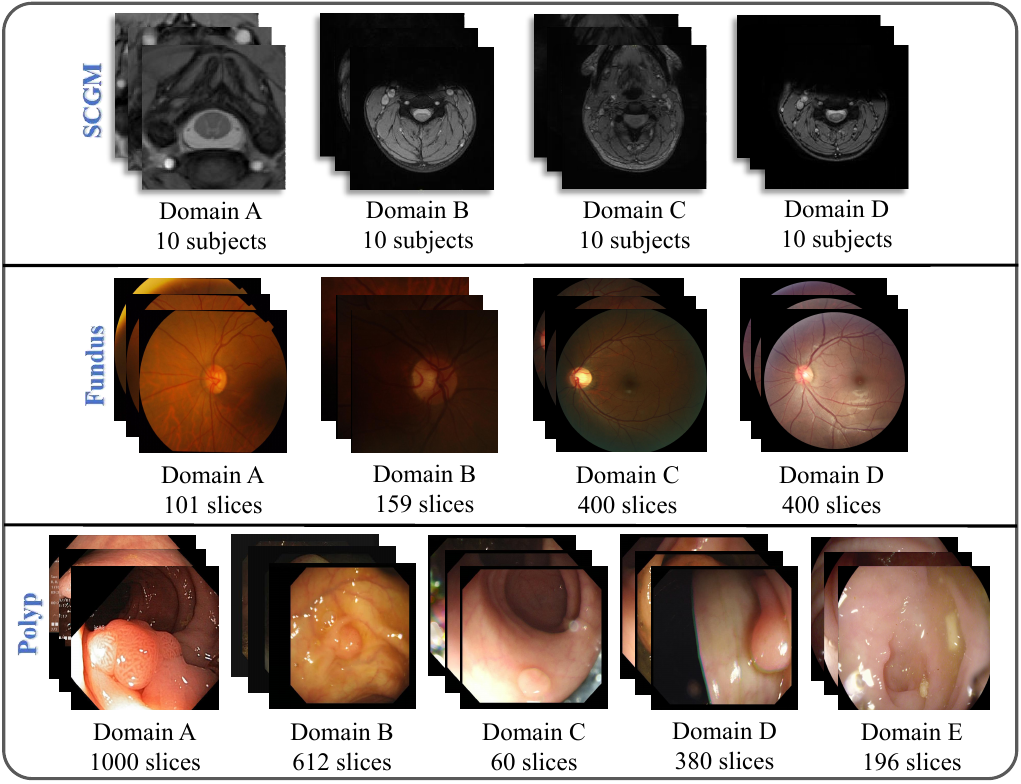}%
   \caption{\textbf{Example cases and sample distribution} for each domain in the \textit{SCGM}, \textit{Fundus} and \textit{Polyp} datasets.}
   \vspace{-.1in}
   \label{fig:dataset}
\end{figure}

\section{Experiments}
\label{sec:experiment}
\noindent
\textbf{Datasets.} We evaluate our DAC on three real-world medical image segmentation datasets: 1) \textbf{Fundus} is a retinal fundus dataset comprising 4 public datasets with different clinical sites or scanners used for optic \textit{disc} and optic \textit{cup} segmentation. 2) \textbf{Polyp} comprises 5 public polyp datasets, namely, Kvasir-SEG \cite{jha2020kvasir}, CVC-ClinicDB \cite{bernal2015wm}, CVC-300 \cite{vazquez2017benchmark}, ColonDB \cite{bernal2012towards} and ETIS \cite{silva2014toward}. 3) \textbf{SCGM} is a spinal cord gray matter segmentation dataset collected from 4 different medical centers with different MRI systems. The voxel resolutions range from $0.25 \times 0.25 \times 0.25$ mm to $0.5 \times 0.5 \times 0.5$ mm. Fig.~\ref{fig:dataset} shows examples and domain sizes for each dataset.

\begin{figure}
  \centering
   \includegraphics[width=1.0\linewidth]{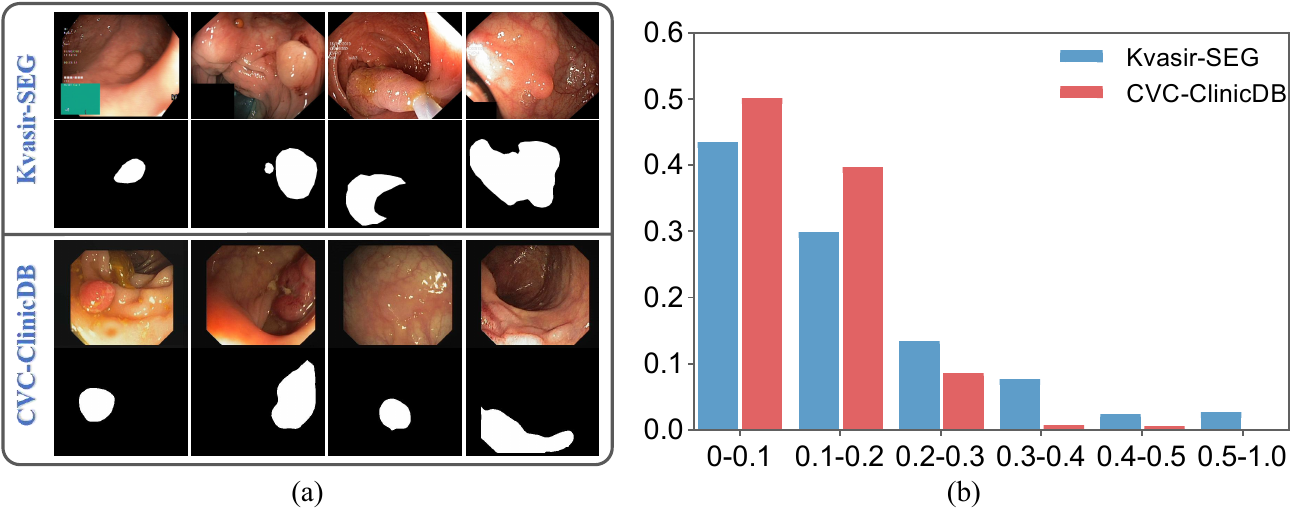}%
   \caption{Examples of \textbf{Kvasir-SEG (domain A)} and \textbf{CVC-ClinicDB (domain B)} from the \textit{Polyp} dataset. (a) Example cases. (b) Distributions of lesion sizes. The horizontal axis represents the proportion of the entire image occupied by the lesion area, whereas the vertical axis denotes the relative frequency of samples exhibiting a specific lesion size in comparison to the entire dataset.}
   \label{fig:AB_size}
\end{figure}

\noindent
\textbf{Setup.} 
To adapt the dataset to the CD-SSDG setting, we designate one domain within a dataset as the target domain (TD), while the remaining domains serve as multiple source domains. We then select a single domain of multiple source domains as the labeled source domain (LSD).
We randomly select a ratio (\textit{e.g.}, 10\% and 20\% for the Fundus dataset) of images from LSD as labeled data, with all remaining images from the source domains treated as unlabeled data. For the Fundus and SCGM datasets, following \cite{zhang2022semi,yao2022enhancing}, we choose one domain as the target and consider the remaining three domains as the sources, which are evaluated on the basis of the average dice scores (DSC).
For the Polyp dataset, Kvasir-SEG (domain A) comprises $1,000$ polyp images along with their respective ground truth annotations generated by expert endoscopists at Oslo University Hospital in Norway \cite{jha2020kvasir}; CVC-ClinicDB (domain B) contains 612 images at a resolution of $384 \times 288$, extracted from 31 colonoscopy sequences conducted at the Hospital Clinic of Barcelona in Spain \cite{bernal2015wm}. As illustrated in Fig.~\ref{fig:AB_size}, polyps have a wide variety of shapes, sizes, colors and appearances. The majority of polyps in both Kvasir-SEG and CVC-ClinicDB are notably diminutive, with the lesion area's occupancy within the entire image predominantly ranging from $0$ to $0.2$.
Following \cite{wu2023acl,fan2020pranet,xianovel} and considering the sizes and characteristics of polyp datasets shown in Fig.~\ref{fig:dataset}, we select training sets from Kvasir-SEG and CVC-ClinicDB (\textit{i.e.}, 900 of domain A and 550 of domain B) as source domains, with the remaining three domains serving as targets. Evaluations consider both the DSC and mean intersection over union (IoU). 
We also evaluate our DAC on the Kvasir-SEG and CVC-ClinicDB testing sets following the approach for ACL-Net \cite{wu2023acl}.

\noindent
\textbf{Implementation details.}
The DAC framework is implemented in PyTorch and executed on two NVIDIA 3090 GPUs.
A ResNet50 model \cite{he2016deep} pretrained on ImageNet \cite{deng2009imagenet} is used as a feature extractor, and DeepLabv3+ \cite{chen2019rethinking} is randomly initialized as a segmentation head for the two sub-models, following prior work \cite{yao2022enhancing,chen2021semi}.
All the models are trained via the AdamW optimizer with an initial learning rate of 0.0001 and a weight decay of 0.01.
We empirically set $\alpha = 0.6, \tau_1 = 0.2, \tau_2 = 0.8$. We itemize the crop size ($cs$), batch size ($bs$), and training epochs ($ep$) for each dataset: 1) Fundus dataset: $cs = 384 \times 384$, $bs = 8$, $ep = 30$, $\lambda_{cps} = 1$, $\lambda_{cfs} = 0.05$, and $\lambda_{ac} = 0.01$. 2) Polyp dataset: $cs = 384 \times 384$, $bs = 8$, $ep = 50$, $\lambda_{cps} = 0.5$, $\lambda_{cfs} = 0.05$, and $\lambda_{ac} = 0.05$. 3) SCGM dataset: $cs = 288 \times 288$, $bs = 8$, $ep = 80$, $\lambda_{cps} = 0.5$, $\lambda_{cfs} = 0.05$, and $\lambda_{ac} = 0.05$. 
The hyper-parameters are chosen after conducting a grid search, \textit{i.e.}, $\lambda_{cps} = [0.1,0.5,1.0,1.5,2.0]$, $\lambda_{cfs} = [0.005,0.010,0.050,0.100,0.200]$, and $\lambda_{ac} = [0.001,0.005,0.010,0.050,0.100]$.
The average of the two sub-model predictions is used for evaluation.
For each configuration involving different source domains and label ratios in each dataset, we conduct experiments with \textit{three} times and report the mean and standard deviation\footnote{Note that we only report the standard deviation for the average performance (average in Table~\ref{tab:table1} \& \ref{tab:table2} \& \ref{tab:table3}) to save space. The standard deviations of all the results can be found in the supplementary material.} of the performance.

\begin{table*}
  \centering
  \caption{\small{
  DSC (\%) results on the \textit{FUNDUS} dataset using 10\% and 20\% labeled data from one source domain, respectively, under the CD-SSDG setting. `Ratio' represents the labeled ratio from the corresponding single-source domain.
  }}
  \begin{tabular}{c||c|c|*{3}{c@{\hspace{7pt}}}|*{3}{c@{\hspace{7pt}}}|*{3}{c@{\hspace{7pt}}}|*{3}{c@{\hspace{7pt}}}||c}
    \toprule[0.5pt]
    \multirow{2}{*}{Method} & \multirow{2}{*}{Ratio} & TD: & \multicolumn{3}{c|}{A} & \multicolumn{3}{c|}{B} & \multicolumn{3}{c|}{C} & \multicolumn{3}{c||}{D} & \multirow{2}{*}{Average} \\
    \cline{3-15}
    &&LSD:& B & C & D & A & C & D & A & B & D & A & B & C & \\   
    \midrule   
    BCP \cite{bai2023bidirectional}& \multirow{8}{*}{10\%} & \multirow{4}{*}{Cup} & 70.98 & 79.62 & 75.64 & 56.61 & 62.46 & 58.60 & 80.09 & 69.62 & \textbf{77.32} & 34.46 & 57.66  & 69.36 & 66.04$\pm$2.01\\
    SOFR \cite{zhang2022semi} &&&61.42&78.68&46.91&53.87&64.01&56.43&67.98&74.29&64.71&58.56&\textbf{68.89}&69.94&63.81$\pm$0.58\\    
    EPL \cite{yao2022enhancing} &&& 75.21&85.93&45.38&71.30 &71.96&52.15&76.73 &70.89&74.58&31.19 &20.68&\textbf{84.57}&63.38$\pm$2.37 \\
    \cellcolor{gray}\textbf{Ours} &&&  \cellcolor{gray}\textbf{76.36} &  \cellcolor{gray}\textbf{86.08} &  \cellcolor{gray}\textbf{72.19} &  \cellcolor{gray}\textbf{76.09} &  \cellcolor{gray}\textbf{71.89} &  \cellcolor{gray}\textbf{71.65} &  \cellcolor{gray}\textbf{83.58} &  \cellcolor{gray}\textbf{76.04} &  \cellcolor{gray}73.07 &  \cellcolor{gray}\textbf{80.21} &  \cellcolor{gray}67.23 &  \cellcolor{gray}81.85 &  \cellcolor{gray}\textbf{76.35$\pm$0.93} \\
    \cline{1-1} \cline{3-16}
    BCP \cite{bai2023bidirectional}&& \multirow{4}{*}{Disc} & 88.19 & 92.15 & 92.20 & 70.98 & 69.67 & 75.52 & 91.50 & 86.53 & 89.80 & 65.22 & 75.27 & 90.34 & 82.28$\pm$1.25 \\
    SOFR \cite{zhang2022semi} &&&86.78&92.60&70.99&81.06&85.35&79.57&86.90&84.83&82.76&72.66&76.77&88.23&82.38$\pm$0.42 \\
    EPL \cite{yao2022enhancing} &&& 92.81&\textbf{96.02}&\textbf{96.09}&83.79&84.24&83.26&92.86&85.77&92.20&\textbf{92.89}&80.71&\textbf{93.77}&89.53$\pm$0.21\\
    
     \cellcolor{gray}\textbf{Ours} &&&  \cellcolor{gray}\textbf{94.19} &  \cellcolor{gray}95.14 &  \cellcolor{gray}94.93 &  \cellcolor{gray}\textbf{85.02} &  \cellcolor{gray}\textbf{85.61} &  \cellcolor{gray}\textbf{86.51} &  \cellcolor{gray}\textbf{93.22} &  \cellcolor{gray}\textbf{87.56} &  \cellcolor{gray}\textbf{94.91} &  \cellcolor{gray}87.21 &  \cellcolor{gray}\textbf{85.06} &  \cellcolor{gray}92.69 &  \cellcolor{gray}\textbf{90.17$\pm$0.33} \\
    \hline \hline
    BCP \cite{bai2023bidirectional} & \multirow{8}{*}{20\%} & \multirow{4}{*}{Cup} &  68.62 & 77.84 & 76.88 & 52.38 & 59.37 & 56.51 & 76.88 & \textbf{76.16} & 78.93 & 31.84 & 62.08 & 62.37 & 64.99$\pm$2.59 \\
    SOFR \cite{zhang2022semi} &&& 62.52 & 78.34 & 46.26 & 58.23 & 68.40 & 58.46 & 69.86 & 68.10 & 68.11 & 50.29 & \textbf{68.04} &  74.44 & 64.25$\pm$1.59\\
    EPL \cite{yao2022enhancing} &&& 79.80 &\textbf{85.70}&65.46&71.52&72.77&58.99& 74.66 &70.16&76.21&44.91&34.14&\textbf{85.39}&68.31$\pm$0.76\\
    \cellcolor{gray}\textbf{Ours} &&& \cellcolor{gray}\textbf{82.12}&\cellcolor{gray}85.17&\cellcolor{gray}\textbf{83.15}&\cellcolor{gray}\textbf{79.04}&\cellcolor{gray}\textbf{78.79}&\cellcolor{gray}\textbf{77.39}&\cellcolor{gray}\textbf{81.02}&\cellcolor{gray}75.36&\cellcolor{gray}\textbf{80.04}&\cellcolor{gray}\textbf{80.85}&\cellcolor{gray}61.91&\cellcolor{gray}83.85&\cellcolor{gray}\textbf{79.06$\pm$0.49}\\
    \cline{1-1} \cline{3-16}
    BCP \cite{bai2023bidirectional} & &\multirow{4}{*}{Disc}& 89.39 & 93.53 & 92.88 & 72.68 & 70.87 & 71.04 & 92.36 & 86.24 & 88.42 & 60.53  & 75.69 & 84.29 & 81.49$\pm$3.51 \\
    SOFR \cite{zhang2022semi} &&& 87.70 & 92.16 & 68.18 & 82.19 & 85.83 & 77.38 & 89.17 & 84.45 & 84.04 & 71.91 & 81.09 & 90.85 & 82.91$\pm$0.45 \\
    EPL \cite{yao2022enhancing} &&&92.77&\textbf{95.82}&95.62&86.52&83.92&84.57&\textbf{93.39}&\textbf{86.83}&93.02&\textbf{93.60}&81.29&\textbf{93.98}&90.11$\pm$0.39\\
    \cellcolor{gray}\textbf{Ours} &&& \cellcolor{gray}\textbf{93.32}&\cellcolor{gray}94.92&\cellcolor{gray}\textbf{95.74}&\cellcolor{gray}\textbf{88.53}&\cellcolor{gray}\textbf{85.95}&\cellcolor{gray}\textbf{87.28}&\cellcolor{gray}92.60&\cellcolor{gray}86.14&\cellcolor{gray}\textbf{94.97}&\cellcolor{gray}90.76&\cellcolor{gray}\textbf{84.55}&\cellcolor{gray}93.68&\cellcolor{gray}\textbf{90.70$\pm$0.24}\\
    \bottomrule[0.5pt]
  \end{tabular}
  \label{tab:table1}
\end{table*}

\noindent
\textbf{Baselines.}
We compare our DAC with \textit{four} state-of-the-art (SOTA) medical image segmentation methods under the CD-SSDG setting. These methods can be categorized into three groups: 
1) \textbf{BCP \cite{bai2023bidirectional}} and \textbf{ACL-Net \cite{wu2023acl}} are designed for semi-supervised segmentation tasks. \textbf{ACL-Net \cite{wu2023acl}} is a SOTA method recognized for semi-supervised polyp segmentation. Notably, \textbf{BCP \cite{bai2023bidirectional}} addresses domain shifts between labeled and unlabeled data with specific design considerations. 
2) \textbf{EPL \cite{yao2022enhancing}} is tailored for the SSDG setting, utilizing a cross pseudo supervision-based co-training framework. Its configuration shares similarties with our Variant \MakeUppercase{\romannumeral1} in Table~\ref{tab:ablation}.
3) \textbf{SOFR \cite{zhang2022semi}} is the only existing method designed for CD-SSDG in medical image segmentation, to the best of our knowledge. Notably, SOFR's CD-SSDG setting assumes more labeled data than our method does, as it typically utilizes an entire source domain as labeled data. In our low-labeled ratio settings (\textit{i.e.}, 10\% and 20\%, as shown in Table~\ref{tab:table1}, 25\% and 50\%, as shown in Table~\ref{tab:table2}, and 20\%, as shown in Table~\ref{tab:table3}), SOFR appears to face challenges in learning generalizable features, resulting in less satisfactory segmentation results for intricate regions such as the optic cup.


\subsection{Comparison Results}

\begin{figure}
  \centering
   \includegraphics[width=1.0\linewidth]{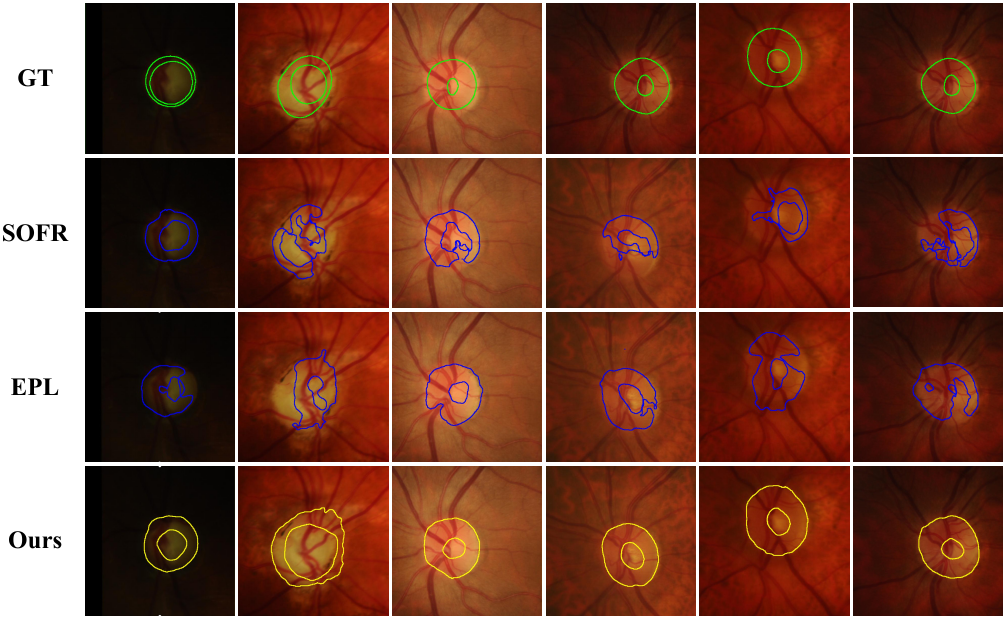}
   \caption{Qualitative comparison results for the \textit{target domain} on the \textit{Fundus} dataset.}
   \label{fig:Q_C}
\end{figure}

\begin{table*}[t]
  \centering
  \caption{\small{DSC (\%) and IoU (\%) results on the \textit{POLYP} dataset using 25\% and 50\% labeled data from one source domain, respectively, under the CD-SSDG setting.
   The third column represents LSD.
   Note that the semi-supervised segmentation method ACL-Net utilizes labeled data from the training sets of the two source domains (A and B) and evaluates the results on the corresponding test sets. While ACL-Net does not have specific designs for SSDG, it provides results for domains C, D, and E as additional showcases for the 25\% labeled ratio configuration. We present its reported results as our upper bound for the comparison on the test sets of domains A and B.
  }}
  \begin{tabular}{c||c|c|*{2}{c@{\hspace{7pt}}}|*{2}{c@{\hspace{7pt}}}|*{2}{c@{\hspace{7pt}}}|*{2}{c@{\hspace{7pt}}}|*{2}{c@{\hspace{7pt}}}||*{2}{c@{\hspace{7pt}}}}
    \toprule[0.5pt]
    \multirow{2}{*}{Method} & \multirow{2}{*}{Ratio}&  TD: &\multicolumn{2}{c|}{A} & \multicolumn{2}{c|}{B} & \multicolumn{2}{c|}{C} & \multicolumn{2}{c|}{D} & \multicolumn{2}{c||}{E} & \multicolumn{2}{c}
    {Average}\\
    
    \cline{3-15}
     &  & LSD & DSC & IoU & DSC & IoU & DSC & IoU & DSC & IoU & DSC & IoU & DSC & IoU  \\
    \midrule
    ACL-Net \cite{wu2023acl}& 25\% & AB & 85.89 & 78.51 & 84.82 & 78.67 & 86.47 & 79.54 & 68.71 & 62.65 & 58.89 & 52.44 & 76.96 & 70.36\\
    \hline
    BCP \cite{bai2023bidirectional} &\multirow{8}{*}{25\%} & \multirow{4}{*}{A} & 79.59 & 70.70 & 68.29 & 59.17 & 64.49 & 54.39 & 56.23 & 46.49 & 42.07 & 34.16& 62.13$\pm$2.28 & 52.98$\pm$2.60 \\
    SOFR \cite{zhang2022semi} &&& 79.18 & 70.55 & 65.81 & 56.32 & 78.50 & 69.25 & 59.68  & 49.96 & 47.00 & 39.22 & 66.03$\pm$2.36 & 57.06$\pm$2.31 \\
    EPL \cite{yao2022enhancing} &&& 87.49 & 81.11 & 75.82 & 68.31 & 87.67 & 79.71 & 66.15 & 58.24 & 62.89 & 54.67 & 76.00$\pm$0.18 & 68.41$\pm$0.25\\     
    \cellcolor{gray}\textbf{Ours} &&& \cellcolor{gray}\textbf{87.65}&\cellcolor{gray}\textbf{81.85}&\cellcolor{gray}\textbf{80.53}&\cellcolor{gray}\textbf{73.90}&\cellcolor{gray}\textbf{89.27}&\cellcolor{gray}\textbf{81.92}&\cellcolor{gray}\textbf{71.58}&\cellcolor{gray}\textbf{64.16}&\cellcolor{gray}\textbf{65.32}&\cellcolor{gray}\textbf{58.37}&\cellcolor{gray}\textbf{78.87$\pm$0.60}&\cellcolor{gray}\textbf{72.04$\pm$0.69}\\
    \cline{1-1} \cline{3-15}
    BCP \cite{bai2023bidirectional}&& \multirow{4}{*}{B} & 73.35 & 62.59 & 74.50 & 66.29 & 57.33 & 45.97  & 48.53 & 39.16 & 39.12 & 30.14 & 58.56$\pm$3.68 & 48.83$\pm$3.73 \\
    SOFR \cite{zhang2022semi} &&& 72.43 & 61.92 & 72.79 & 64.47 & 70.32 & 59.93 & 54.82 & 45.42 & 39.15 & 30.30 & 61.90$\pm$2.85& 52.41$\pm$3.01\\
    EPL \cite{yao2022enhancing} &&& 82.10 & 74.19 & 82.59 & 75.51 & 88.29 & 80.82 & 66.77 & 58.54 & 59.55 & 50.74 & 75.86$\pm$1.33& 67.96$\pm$1.23 \\
    \cellcolor{gray}\textbf{Ours} &&& \cellcolor{gray}\textbf{86.18}&\cellcolor{gray}\textbf{78.60}&\cellcolor{gray}\textbf{83.21}&\cellcolor{gray}\textbf{77.09}&\cellcolor{gray}\textbf{90.22}&\cellcolor{gray}\textbf{83.12}&\cellcolor{gray}\textbf{69.32}&\cellcolor{gray}\textbf{62.16}&\cellcolor{gray}\textbf{65.99}&\cellcolor{gray}\textbf{58.85}&\cellcolor{gray}\textbf{78.98$\pm$0.98}&\cellcolor{gray}\textbf{71.96$\pm$0.83}\\
    \hline \hline
    ACL-Net \cite{wu2023acl}& 50\% & AB & 88.12 & 81.74 & 87.25 & 81.64 & - & - & - & - & - & - & - & - \\
    \hline
    BCP \cite{bai2023bidirectional} & \multirow{8}{*}{50\%} & \multirow{4}{*}{A} & 81.52 & 73.23 & 74.77 & 65.83 & 64.71 & 54.81 & 59.44 & 49.71 & 44.09 & 35.97 & 64.91$\pm$2.97 & 55.91$\pm$3.24\\
    SOFR \cite{zhang2022semi} &&& 82.05 & 73.87 & 70.00 & 61.25 & 77.84 & 67.77 & 64.07 & 54.64 & 49.26 & 41.80 & 68.64$\pm$1.51 & 59.87$\pm$1.86 \\
    EPL \cite{yao2022enhancing} &&& 88.08 & 81.92 & 78.02 & 70.86 & 87.95 & 80.30 & 67.46 & 59.58 & 63.34 & 55.83 & 76.97$\pm$0.67 & 69.70$\pm$0.50  \\ 
    \cellcolor{gray}\textbf{Ours} & & &\cellcolor{gray}\textbf{88.18}&\cellcolor{gray}\textbf{82.55}&	\cellcolor{gray}\textbf{81.82}&\cellcolor{gray}\textbf{75.25}&\cellcolor{gray}\textbf{88.76}&\cellcolor{gray}\textbf{81.11}&\cellcolor{gray}\textbf{71.08}&\cellcolor{gray}\textbf{64.05}&\cellcolor{gray}\textbf{68.53}&\cellcolor{gray}\textbf{61.48}&\cellcolor{gray}\textbf{79.67$\pm$0.52}&\cellcolor{gray}\textbf{72.89$\pm$0.45}\\
    \cline{1-1} \cline{3-15}
    BCP \cite{bai2023bidirectional} && \multirow{4}{*}{B} & 73.37 & 62.30 & 76.59 & 68.57 & 55.27 & 43.90 & 49.46 & 39.72 & 40.97 & 31.30 & 59.13$\pm$2.44 & 49.16$\pm$2.71 \\
    SOFR \cite{zhang2022semi} &&& 76.15 & 66.63 & 78.35 & 70.66 & 77.29 & 68.33 & 62.20 & 53.26 & 50.44 & 41.16 & 68.89$\pm$1.08 & 60.01$\pm$1.20\\
    EPL \cite{yao2022enhancing} &&& 80.29 & 72.44 & 83.81 & 77.00 & 88.21 & 80.88 & 67.89 & 59.57 & 60.08 & 51.50 & 76.06$\pm$1.24 & 68.28$\pm$1.25\\

    \cellcolor{gray}\textbf{Ours} &&&\cellcolor{gray}\textbf{85.64}&\cellcolor{gray}\textbf{78.24}&\cellcolor{gray}\textbf{85.95}&\cellcolor{gray}\textbf{79.97}&\cellcolor{gray}\textbf{88.60}&\cellcolor{gray}\textbf{81.34}&\cellcolor{gray}\textbf{69.06}&\cellcolor{gray}\textbf{61.71}&\cellcolor{gray}\textbf{66.26}&\cellcolor{gray}\textbf{58.91}&\cellcolor{gray}\textbf{79.10$\pm$0.92}&\cellcolor{gray}\textbf{72.03$\pm$0.77}\\
    \bottomrule[0.5pt]
  \end{tabular}
  \label{tab:table2}
\end{table*}

\begin{table*}
  \centering
  \caption{\small{DSC (\%) results on the \textit{SCGM} dataset using 20\% labeled data from one source domain, respectively, under the CD-SSDG setting. 
  }}
  \begin{tabular}{c||c|ccc|ccc|ccc|ccc||c}
    \toprule[0.5pt]
    \multirow{2}{*}{Method} & TD: & \multicolumn{3}{c|}{A} & \multicolumn{3}{c|}{B} & \multicolumn{3}{c|}{C} & \multicolumn{3}{c||}{D} & \multirow{2}{*}{Average} \\
    \cline{2-14}
    &LSD:& B & C & D & A & C & D & A & B & D & A & B & C & \\   
    \midrule   
    BCP \cite{bai2023bidirectional}&\multirow{4}{*}{\makecell{20\% \\ labeled}} &81.82&82.59&83.14&48.71&80.92&86.12&49.14&52.13&55.46&75.13&	90.24&87.58&72.75$\pm$ 3.87\\
    SOFR \cite{zhang2022semi}&&56.67&68.78&63.56&55.51&78.52&86.75&28.09&43.08&44.39&60.37&82.39&76.21&	62.03$\pm$1.60\\ 
    EPL \cite{yao2022enhancing} && 82.76&82.10&\textbf{84.43}&50.87&81.64&88.98&69.37&71.80&74.30&30.92&	89.55&88.22&74.58$\pm$3.39\\
    \cellcolor{gray}\textbf{Ours} &&\cellcolor{gray}\textbf{83.09}&\cellcolor{gray}\textbf{84.13}&\cellcolor{gray}84.37&\cellcolor{gray}\textbf{80.87}&\cellcolor{gray}\textbf{84.86}&\cellcolor{gray}\textbf{89.24}&\cellcolor{gray}\textbf{71.24}&\cellcolor{gray}\textbf{75.62}&\cellcolor{gray}\textbf{76.43}&\cellcolor{gray}\textbf{80.32}&	\cellcolor{gray}\textbf{90.46}&\cellcolor{gray}\textbf{89.43}&\cellcolor{gray}\textbf{82.51$\pm$0.67} \\
    \bottomrule[0.5pt]
  \end{tabular}
  \label{tab:table3}
\end{table*}

\begin{figure}
  \centering
   \includegraphics[width=1.0\linewidth]{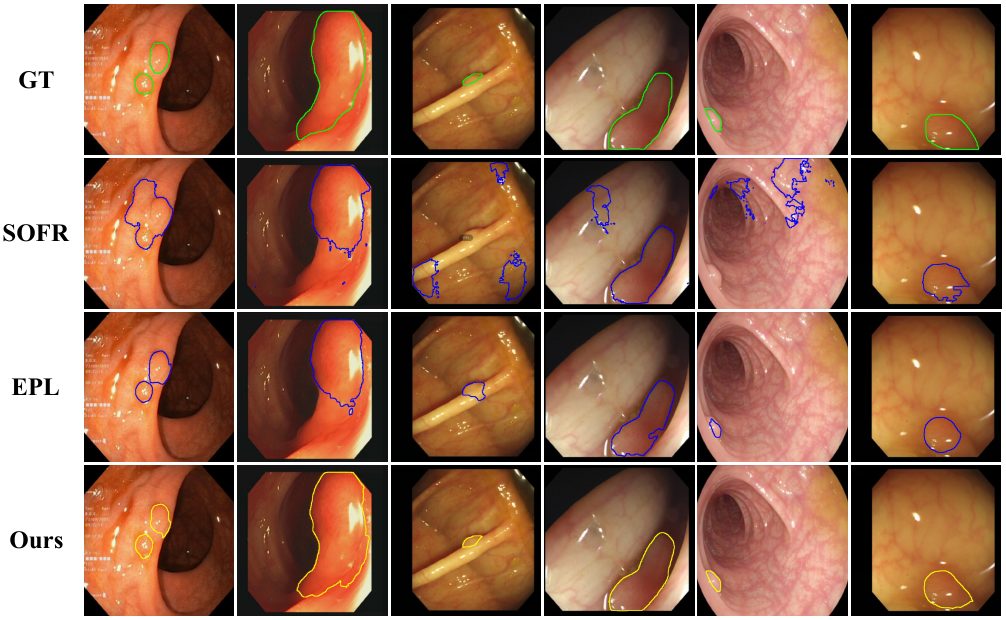}
   \caption{Qualitative comparison results for the \textit{target domain} on the \textit{Polyp} dataset.}
   \label{fig:Q_C_polyp}
\end{figure}

\begin{figure}
  \centering
   \includegraphics[width=1.0\linewidth]{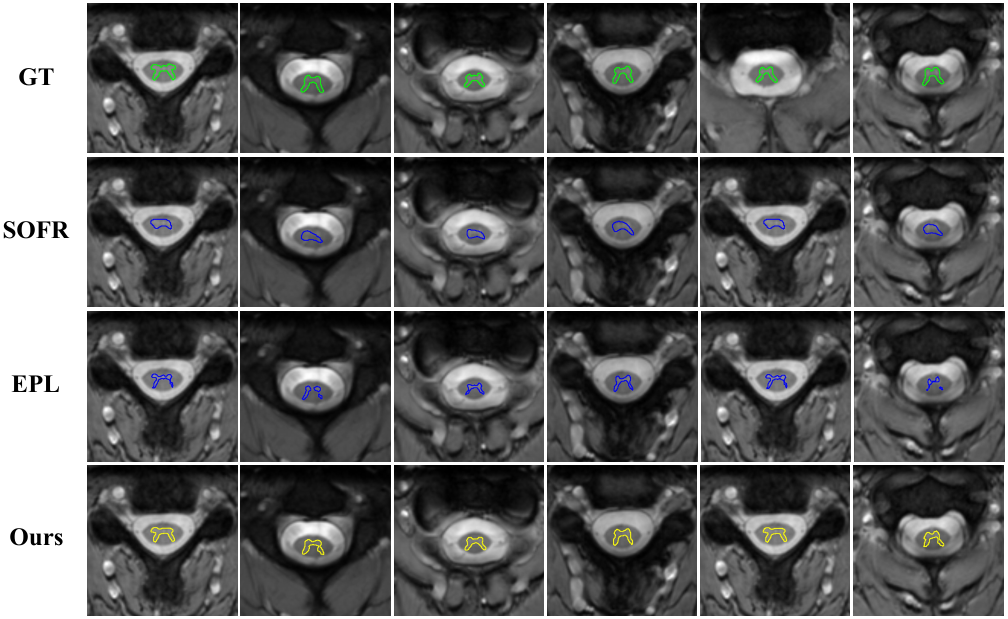}
   \caption{Qualitative comparison results for the \textit{target domain} on the \textit{SCGM} dataset.}
   \label{fig:Q_C_scgm}
\end{figure}

\noindent
\textbf{Results on the Fundus dataset.} 
Table~\ref{tab:table1} presents the comparative results for retinal fundus segmentation under the CD-SSDG scenario, considering labeled ratios of 10\% and 20\%. Our DAC method outperforms baseline approaches, achieving the best overall generalization performance (Average) and demonstrating improvements across most configurations for both optic disc and cup segmentation. Notably, in the more challenging optic cup segmentation task, compared with the best results from the baselines with 20\% and 10\% labeled data, we observe an increase in the average DSC score by $+10.75\%$ (vs. EPL) and $+10.31\%$ (vs. BCP), respectively. These notable improvements in challenging tiny cup segmentation underscore the effectiveness of our dual-supervised asymmetric co-training framework. The synergy of dual-supervision and asymmetric co-training successfully learns domain-generalizable and discriminative features, enabling accurate segmentation of edge regions and low-contrast areas in images from previously unseen domains. 
Additionally, we present qualitative results comparing the DAC with SSDG methods (\textit{i.e.}, EPL and SOFR) in Fig.~\ref{fig:Q_C}. Notably, DAC is less sensitive to the vascularity of object boundaries, producing more coherent segmentation results across the six examples from the unseen domain. This observation highlights the strong generalizability of DAC in fine-grained segmentation.

\noindent
\textbf{Results on Polyp dataset.} Table~\ref{tab:table2} reports the comparison results for polyp segmentation, considering ratios of 25\% and 50\%. 
Our DAC consistently outperforms other methods across various configurations under the CD-SSDG setting. These results highlight the superior generalizability of our method compared with that of various baseline groups, including those with specific designs for CD-SSDG (\textit{i.e.}, SOFR). 
Notably, our DAC shows competitive performance (\textit{i.e.}, tested on domain B with a 25\% labeled ratio) or even outperforms the performance (\textit{i.e.}, tested on domain A with a 25\% labeled ratio) compared with that of the upper bound ACL-Net.
We present qualitative results for polyp segmentation in Fig.~\ref{fig:Q_C_polyp}. Across six examples from the unseen domain in this challenging dataset, the targeted segmentation region has varying sizes and shapes. 
Even for challenging objects with a small size or off-center position, our DAC achieves impressive segmentation results that closely align with the ground truth.

\noindent
\textbf{Results on the SCGM dataset.} 
Table~\ref{tab:table3} shows the comparative results for spinal cord gray matter (SCGM) segmentation under the CD-SSDG scenario, where our method results in an improvement of $+7.93\%$ in terms of the DSC score with $20\%$ labeled data compared with that of the previous best method (\textit{i.e.}, EPL). Furthermore, Fig.~\ref{fig:Q_C_scgm} presents the qualitative results for SCGM segmentation. The results clearly demonstrate that our DAC is able to accurately segment unknown distributions. 
Moreover, our method yields segmentation results with sharper and more coherent object boundaries; as a result, we attribute its effectiveness in mitigating the detrimental effects of inaccurate pseudo-labels and uncovering discriminative and diverse features.

\begin{table}[t]
  \centering
  \caption{Ablation study on alternative designs of DAC.}
  \scalebox{0.85}{
  \begin{tabular}{c|c|*{3}{cc|}c}
    \toprule
    \multirow{2}{*}{Variant}&\multirow{2}{*}{CFS} & \multicolumn{2}{c|}{sub-model 1} & \multicolumn{2}{c|}{sub-model 2}  & \multirow{2}{*}{Cup}  & \multirow{2}{*}{Disc} & \multirow{2}{*}{Average}  \\
    &&$H_{loc}$ &$H_{rot}$ & $H_{loc}$ &$H_{rot}$ &&& \\
    \midrule
    \MakeUppercase{\romannumeral1}&-&-&-&-&-&68.48&86.94&77.71 \\
    \MakeUppercase{\romannumeral2}&-&\checkmark&-&-&\checkmark&76.51&86.44&81.47 \\
    \MakeUppercase{\romannumeral3}&\checkmark &-&-&-&-&74.57&86.83&80.70\\\midrule
    \MakeUppercase{\romannumeral4}&\checkmark & \checkmark &-& \checkmark &-&76.48&87.61& 82.04 \\
    \MakeUppercase{\romannumeral5}&\checkmark &-&\checkmark&-&\checkmark&75.87&86.77&81.32\\     \midrule
    \textbf{Ours}
    &\checkmark &\checkmark&-&-&\checkmark&\textbf{79.11}&\textbf{87.70}&\textbf{83.40} \\
    \bottomrule
  \end{tabular}}
  \label{tab:ablation}
\end{table}

\begin{figure}[t]
  \centering
   \includegraphics[width=0.70\linewidth]{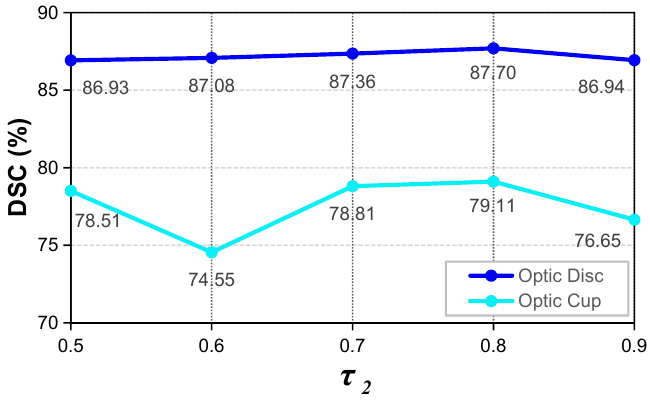}%
   \caption{Impact of $\tau_2$ on the CutMix region size.}
   \label{fig:abla_cutmix}
\end{figure}

\begin{figure}[t]
  \centering
   \includegraphics[width=1\linewidth]{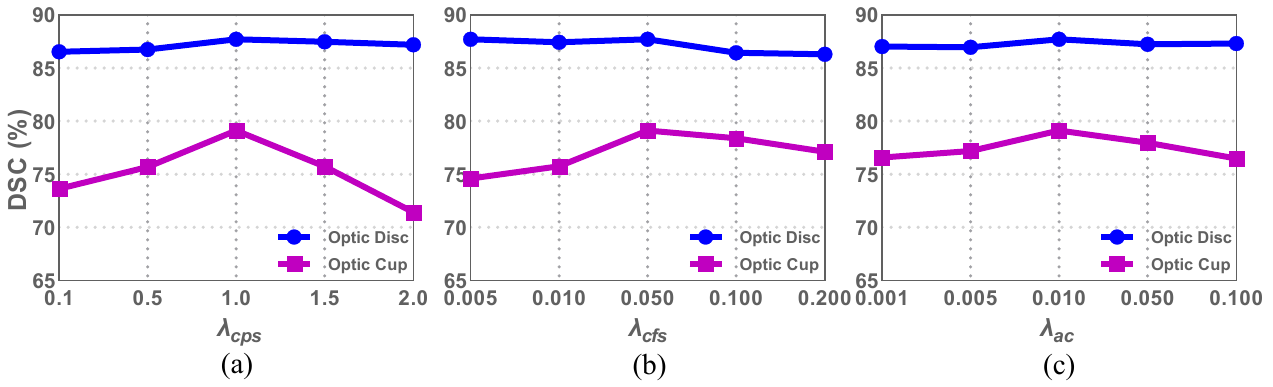}%
   \caption{\small{
   Impact of $\lambda_{cps}$,  $\lambda_{cfs}$ and $\lambda_{ac}$.}
   }
   \label{fig:lam}
\end{figure}

\begin{figure}[t]
\centering
\subfloat[Optic Cup]{\includegraphics[scale=0.55]{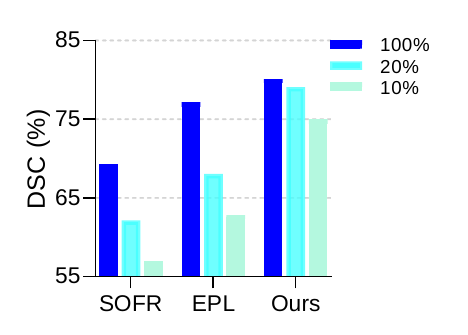}
\label{fig:subfig:Ratio_cup}}
\hfil
\subfloat[Optic Disc]{\includegraphics[scale=0.55]{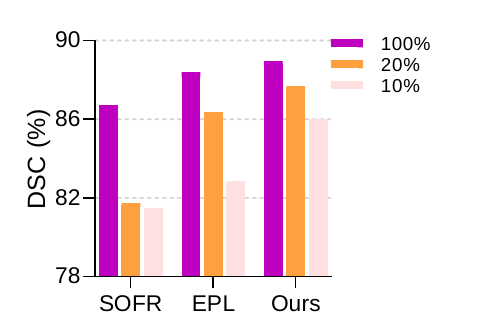}%
\label{fig:subfig:Ratio_disc}}
\caption{Comparison of different labeled ratios with only the source domain.}
\label{fig:Ratio}
\end{figure}

\begin{table}[t]
  \centering
  \caption{
  Time-Consuming comparison.}
  {
  \begin{tabular}{c|c|c}
    \toprule
    Method& Training time (mins) & Inference time (ms)\\
    \midrule
    BCP&498 &54.0 \\
    SOFR&126 &42.5 \\
    EPL&96 &60.1 \\
    \textbf{Ours} &83 &57.3\\
    \bottomrule
  \end{tabular}}
  \label{tab:time}
\end{table}

\begin{figure}[t]
  \centering
   \includegraphics[width=0.95\linewidth]{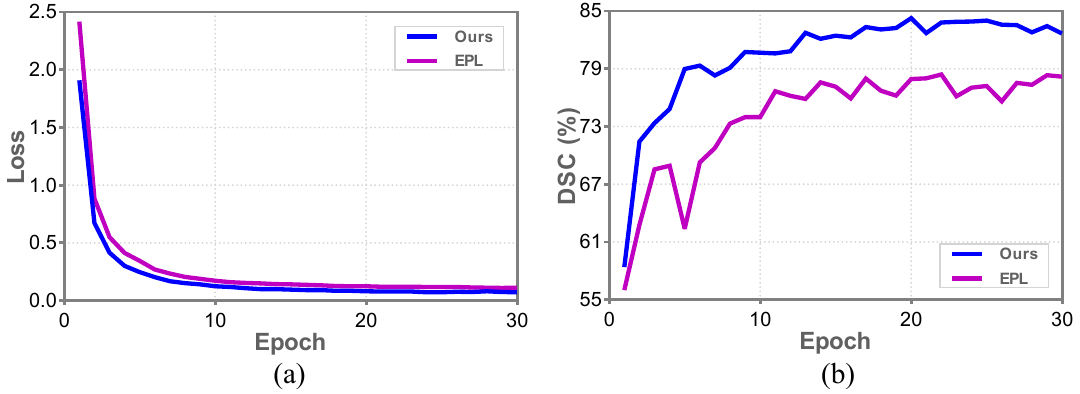}
   \caption{Comparison of the training loss curve (a) and the average performance on the test dataset (b) for the optic cup and disc between our method and EPL.}
   \label{fig:stable}
\end{figure}

\subsection{Ablation Study}
We conduct an ablation study on the Fundus dataset using 20\% labeled data, with domain B as the target domain and the others as the source domains. We report the average result over three configurations, each utilizing labeled data from one of the three source domains.

\noindent
\textbf{Contribution of Each Component.} To assess the specific contributions of our proposed dual supervision mechanism and asymmetric co-training scheme, we conduct ablation experiments on the Fundus dataset, as detailed in Table~\ref{tab:ablation}. We set the variant using only confidence-aware cross pseudo supervision for co-training as our baseline (Variant \MakeUppercase{\romannumeral1}). The results show that the addition of cross feature supervision (Variant \MakeUppercase{\romannumeral3}) significantly improves the model performance. Furthermore, the introduction of asymmetric co-training (Variant \MakeUppercase{\romannumeral2}) leads to an even greater improvement, supporting our hypothesis that diverse auxiliary self-supervised learning tasks in asymmetric co-training enhance the model's robustness in domain-shifted training environments, making it more resilient to noise in pseudo-labels. Our final model that integrates both dual supervision and asymmetric co-training components yields the best performance.

\noindent
\textbf{Effect of the Asymmetric Co-training Scheme.}
To validate our design choice of the asymmetric co-training scheme, we compare it with a symmetric co-training scheme, where one auxiliary self-supervised task is applied to both sub-models. Specifically, by setting the task as either mixed patch localization ($H_{loc}$) or random patch rotation prediction ($H_{rot}$), we create Variants \MakeUppercase{\romannumeral4} and \MakeUppercase{\romannumeral5}, as shown in Table~\ref{tab:ablation}. A comparison of these two variants with our final model reveals that our asymmetric design achieves better performance. This result suggests that our approach provides a more effective integration of auxiliary self-supervised learning within the co-training paradigm.

\noindent
\textbf{Effect of the CutMix Region Size.}
Additionally, we investigate the impact of the zero-value region size when performing CutMix on the Fundus dataset, as it is related to both data augmentation and the auxiliary self-training task. Since the region size is determined by $\beta \sim U(\tau_1,\tau_2)$, as discussed in Section~\ref{subsec:DL}, and $\tau_1$ cannot be too small to ensure that the mixed patch has a sufficient region, we fix $\tau_1 = 0.2$ and vary $\tau_2$ in the range of $(0.5, 0.9)$. The results are presented in Fig.~\ref{fig:abla_cutmix}. Notably, only the performance on the optic cup is sensitive to the change in $\tau_2$, whereas the performance on the optic disc remains stable. We choose $\tau_2 = 0.8$, which yields the best average performance on both the cup and disc segmentation tasks, as our final setting.

\noindent
\textbf{Impact of Hyper-parameters.}
Figure~\ref{fig:lam} shows the sensitivity of our DAC to hyper-parameters $\lambda_{cps}$, $\lambda_{cfs}$ and $\lambda_{ac}$. Our model exhibits minimal variance in optic disc segmentation across a variety of hyper-parameters, whereas the performance on the optic cup experiences some fluctuations. This outcome could be attributed to the severe class imbalance and low contrast between the optic cup and its background, which poses a challenge for the model in segmenting the optic cup accurately. Overall, the variation in hyper-parameters does not lead to a significant decline in performance, underscoring the stability of our approach.

\noindent
\textbf{Different Percentages of Labeled Data.} Fig.~\ref{fig:Ratio} presents the results for optic cup and disc segmentation, considering labeled ratios of $100\%$, $20\%$, and $10\%$. Our DAC maintains superior performance for both segmentation tasks as the labeled ratio decreases. 
While SOFR, which is also designed for CD-SSDG, exhibits a substantial decline in performance for both tasks, \textit{e.g.,}, the performance of the optic cup decreases by $-12.37\%$ when the labeled ratio decreases from $100\%$ to $10\%$.
This finding indicates that our DAC can mitigate the negative impact of domain-shifted training data and effectively leverage unlabeled data to learn domain-invariant discriminative features, thus achieving high generalization performance.

\subsection{Time Consumption and Stability}
We evaluate the training time on the Fundus dataset in Table~\ref{tab:time}.  
Our DAC demonstrates the most efficient training time. This efficiency stems from the fact that we use original images in a sub-model to generate supervisory signals for another sub-model to obtain a high-quality pseudo-label, whereas EPL uses both original and augmented images. Moreover, auxiliary self-supervised tasks and the feature-level supervision signal help the model acquire more useful information within a single epoch, thereby enhancing the model's generalization performance and reducing the number of training epochs needed. Furthermore, these schemes are implemented only during the training process, ensuring that there is no impact on the inference complexity. Therefore, our DAC matches the inference times of the other baselines.

To investigate the stability of DAC, we plot the training loss and test set performance curves between our DAC and the SSDG method EPL \cite{yao2022enhancing}, which is also based on a co-training framework, on the Fundus dataset.
As delineated in Fig.~\ref{fig:stable}(a), the loss curve clearly demonstrates that our DAC does not suffer from convergence issues, suggesting that the dual supervision and asymmetric co-training components do not necessarily face convergence and stability issues even though they can increase the model complexity. Notably, these two components help the model learn discriminative yet diverse features, which enhances the model's generalizability on the unseen test set, as depicted in Fig.~\ref{fig:stable}(b).

\begin{figure}
  \centering
   \includegraphics[width=1.0\linewidth]{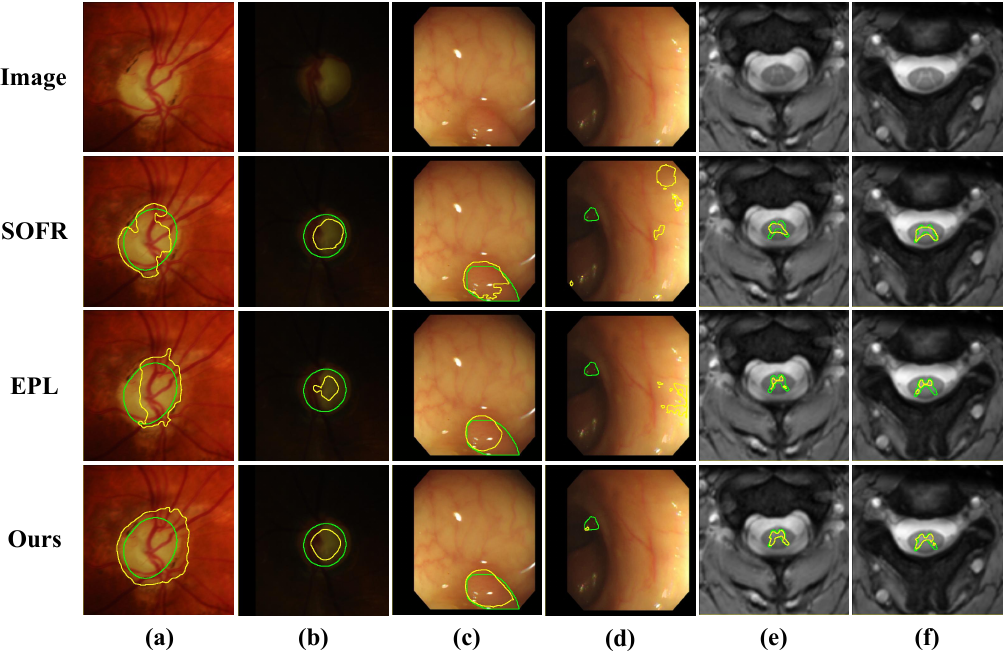}
   \caption{
   {Failure cases. Yellow and green contours denote segmentation results of the methods and the ground truth, respectively.}}
   \label{fig:Failure}
\end{figure}
\subsection{Failure Case Analysis}
We also note a few failure cases. Fig.~\ref{fig:Failure} presents these instances, with two examples each for the Fundus, Polyp, and SCGM datasets. The first instance of failure is depicted in Fig.~\ref{fig:Failure}(a). When multiple blood vessels traverse the optic disc, the variability inherent in these blood vessels induces significant alterations in the optic disc's appearance, thereby undermining the segmentation accuracy of the model. Considering the distinctive characteristics of the Fundus dataset, we propose an augmentation technique that employs a mask or pseudo-label to extract blood vessels from fundus images and mix them with the optic disc to generate additional training data. These augmented images enable the model to learn more diverse features of optic discs with intersecting blood vessels, thereby enhancing the model's efficacy in the unseen domain.
In addition, methods may still encounter difficulties when the color contrast between small objects and the background is extremely low (Fig.~\ref{fig:Failure}(b-f)). The low contrast, in conjunction with the severe class imbalance issue, poses a challenge for the feature extractor to distinguish the object from the background and to capture its features effectively. Inspired by \cite{nam2024modality,xu2021fourier}, we extract features from multiple views (\textit{e.g.}, scale and frequency), where multiscale analysis captures intricate details and broader structural information, and the frequency domain, which preserves the high-level semantics of the image, aids the model in capturing boundary features.


\section{Conclusion}
\label{sec:conclusion}

In this paper, we present a novel dual-supervised asymmetric co-training (DAC) framework for CD-SSDG in medical image segmentation. 
Specifically, we introduce a dual-level supervised mechanism that incorporates feature-level supervision to complement cross pseudo supervision. This dual-supervision approach mitigates the negative impacts of inaccurate pseudo-labels caused by the domain-shifted training environment in CD-SSDG.
Furthermore, we design an asymmetric co-training scheme that incorporates two auxiliary self-supervised learning tasks: 
mixed patch localization and random patch rotation prediction. 
This unique scheme not only prevents the co-trained sub-models from collapsing but also facilitates the learning of discriminative and transferable features. Finally, we evaluate DAC on three public benchmark datasets, \textit{i.e.}, Fundus, Polyp, and SCGM. 
Its superior performance compared with that of several SOTA baselines under various configurations demonstrates its effectiveness and generalizability for CD-SSDG in medical image segmentation.

Severe class imbalance and low contrast between the foreground and background are common challenges in medical image segmentation, and they are particularly exacerbated in the context of CD-SSDG. 
To mitigate the performance degradation caused by these issues, future studies could explore an adaptive, multi-view methodology. This approach adaptively adjusts the image scale to capture broader and more detailed structural features. Moreover, it enhances the extraction of boundary features from the frequency view, thereby refining the delineation of object contours and improving the segmentation accuracy.

\bibliographystyle{IEEEtran}
\bibliography{sample-base}

\end{document}